\documentclass[letterpaper]{article} 
\usepackage{aaai2026}  
\usepackage{times}  
\usepackage{helvet}  
\usepackage{courier}  
\usepackage[hyphens]{url}  
\usepackage{graphicx} 
\usepackage{natbib}  
\usepackage{caption} 
\usepackage{algorithm}
\usepackage{algorithmic}

\usepackage{amsmath} 
\usepackage{newfloat}
\usepackage{listings}
\DeclareCaptionStyle{ruled}{labelfont=normalfont,labelsep=colon,strut=off} 
\floatstyle{ruled}
\newfloat{listing}{tb}{lst}{}
\floatname{listing}{Listing}
\title{LLandMark: A Multi-Agent Framework for Landmark-Aware
Multimodal Interactive Video Retrieval\thanks{This research is supported by AI VIETNAM.}}
\author {
    Minh-Chi Phung\textsuperscript{\rm 1,3},
    Thien-Bao Le\textsuperscript{\rm 1,3},
    Cam-Tu Tran-Thi\textsuperscript{\rm 1,3}
    Thu-Dieu Nguyen-Thi\textsuperscript{\rm 1,3}
    Vu-Hung Dao\textsuperscript{\rm 1,3}
    Tinh-Anh Nguyen-Nhu\textsuperscript{\rm 2,3}
}
\affiliations {
    \textsuperscript{\rm 1}University of Information Technology, Ho Chi Minh City, Vietnam\\
    \textsuperscript{\rm 2}Ho Chi Minh University of Technology, Ho Chi Minh City, Vietnam\\
    \textsuperscript{\rm 3}Vietnam National University, Ho Chi Minh City, Vietnam \\
    {\{23520179, 23520110, 23521704, 23520289, 23520554\}}@gm.uit.edu.vn, anh.nguyennhu2306@hcmut.edu.vn
}

\usepackage{bibentry}

\begin{document}

\maketitle

\begin{abstract}
The increasing diversity and scale of video data demand retrieval systems capable of multimodal understanding, adaptive reasoning, and domain-specific knowledge integration. This paper presents LLandMark, a modular multi-agent framework for landmark-aware multimodal video retrieval to handle real-world complex queries. The framework features specialized agents that collaborate across four stages: query parsing and planning, landmark reasoning, multimodal retrieval, and reranked answer synthesis. A key component, the Landmark Knowledge Agent, detects cultural or spatial landmarks and reformulates them into descriptive visual prompts, enhancing CLIP-based semantic matching for Vietnamese scenes. To expand capabilities, we introduce an LLM-assisted image-to-image pipeline, where a large language model (Gemini 2.5 Flash) autonomously detects landmarks, generates image search queries, retrieves representative images, and performs CLIP-based visual similarity matching, removing the need for manual image input. In addition, an OCR refinement module leveraging Gemini and LlamaIndex improves Vietnamese text recognition. Experimental results show that LLandMark achieves adaptive, culturally grounded, and explainable retrieval performance.
\end{abstract}

\section{Introduction}

Retrieving evidence from large, heterogeneous, and weakly structured video corpora presents major challenges due to multilingual variability, noisy text extraction, and the need for robust cross-modal reasoning. Building upon prior benchmarks such as the Lifelog Search Challenge \cite{ref_lsc} and the Video Browser Showdown \cite{ref_vbs}, the Ho Chi Minh City AI Challenge (HCMAIC) 2025 \cite{ref_hcmaic} extends this research frontier by introducing new benchmarks for temporal reasoning and multimodal understanding in large-scale video datasets, emphasizing systems capable of aligning visual and textual information over time and retrieving complex event-based evidence from a 250GB corpus.

Despite progress from prior works, such as MAVEN’s \cite{ref_maven} agent-based framework with limited adaptability, OCR-driven pipelines leveraging DeepSolo \cite{ref_deepsolo} and PARSeq \cite{ref_parseq} that remain computationally heavy, and RAG-based fusion methods \cite{ref_rag} that struggle with scalability, significant opportunities remain for improving efficiency, flexibility, and multilingual reasoning in large-scale video retrieval.

Most existing systems still overlook spatial and cultural context, particularly \textbf{landmark reasoning}, which is crucial for Vietnamese queries (e.g., “in front of St.~Joseph’s Cathedral in Hanoi” or “near Turtle Tower”). To address this, our framework LLandMark introduces a multi-agent architecture for adaptive planning and multimodal reasoning. The \textbf{Parsing and Planning Agent} interprets query intent and determines the optimal search sequence, the \textbf{Landmark Knowledge Agent} reformulates landmark names into visual descriptions, an \textbf{Orchestrator} executes the search plan in parallel, and the \textbf{Reranking and Answer Agent} fuses multimodal results.

We further introduce an \textbf{OCR Refinement Module} that improves Vietnamese text quality by combining PaddleOCR \cite{ref_paddleocr} with Gemini 2.5 Flash via LlamaIndex \cite{ref_llamaindex}. Finally, an \textbf{LLM-assisted image-to-image retrieval pipeline} automatically detects landmarks in user queries, retrieves representative images, and performs CLIP-based similarity matching \cite{ref_clip}, enabling automated and culturally grounded retrieval of Vietnamese landmarks and scenes.


In summary, our key contributions are as follows:
\begin{itemize}
\item \textbf{LLandMark}: A modular multi-agent architecture for query planning, landmark reasoning, and multimodal reranking.
\item \textbf{OCR with Gemini Refinement}: A hybrid OCR correction pipeline combining PaddleOCR and Gemini-based post-processing via LlamaIndex.
\item \textbf{LLM-Assisted Landmark Image-to-Image Retrieval}: Automated pipeline that detects landmarks, retrieves web images, and performs CLIP-based similarity matching.
\end{itemize}

\section{Related Work}
Recent advances in interactive video retrieval leverage multimodal learning \cite{vqa2024,nguyennhu2025stervlmspatiotemporalenhancedreference}, OCR, and agent systems, yet most frameworks \cite{ref_rag,ref_artemis,ref_avisearch,nguyennhu2025lightweightmomentretrievalglobal,tran2025efficientrobustmomentretrieval} focus on isolated components rather than tight modular integration. Benchmark-driven systems like Vibro \cite{ref_vibro2024} and LifeSeeker 6.0 \cite{ref_lifeseeker24} improve interactivity and linguistic alignment but remain primarily user-driven, relying on manual query reformulation instead of autonomous reasoning.

Within HCMAIC, MAVEN \cite{ref_maven} introduced a multi-agent workflow but was limited in planning and landmark reformulation. Other systems \cite{ref_avisearch,ref_mmmsvr,ref_revimm,nguyen2025hybridunifiediterativenovel} relied on fixed, inflexible pipelines. Our work advances this direction with a modular multi-agent framework featuring dedicated agents for parsing, landmark reasoning, and multimodal reranking.

Scene-text retrieval often uses a two-stage pipeline combining detectors like DeepSolo \cite{ref_deepsolo} with recognition models like PARSeq \cite{ref_parseq}, as in \cite{ref_unveil,ref_addr_ambiguous}, creating a computational bottleneck. Instead, our system avoids this costly pre-processing with a Gemini-based post-correction module via LlamaIndex \cite{ref_llamaindex}, refining noisy OCR output and improving Vietnamese text quality for downstream retrieval.

Prior systems support image-to-image retrieval \cite{ref_artemis,ref_mmmsvr,ref_lameframes,ref_enhanced,ref_snapseek} but depend on manual image selection and lack semantic understanding of landmarks or cultural contexts. In contrast, our automated LLM-assisted pipeline uses Gemini 2.5 Flash to detect landmarks, construct refined web-image queries, fetch representative images, and perform CLIP-based matching \cite{ref_clip}, enabling fully autonomous, culturally grounded retrieval of Vietnamese landmarks and scenes.

\section{Preprocessing}

\subsection{Dataset and Tasks.}
Experiments were conducted on the official dataset of the HCMAIC 2025, which provides a large-scale 250~GB corpus of broadcast and documentary videos spanning diverse domains such as news, education, travel, culture, and sports. A significant portion of the dataset consists of news and reportage content, providing rich contextual cues for multimodal retrieval.

The challenge defines three evaluation tasks:
\begin{itemize}
    \item \textbf{Textual-KIS (Keyword-based Information Search):} Retrieval based on textual keywords extracted from speech, on-screen text, and semantic descriptions.
    \item \textbf{Visual Question Answering (QA):} Frame-level reasoning to identify visual evidence supporting answers to natural-language questions.
    \item \textbf{Temporal Reasoning and Keyframe Extraction (TRAKE):} Event-sequence detection requiring retrieval of temporally ordered video segments.
\end{itemize}
All tasks adopt a unified submission format, where each prediction includes the video name and corresponding frame indices.
\subsection{Data Pre-processing.}
Effective offline data pre-processing forms the foundation of a robust video retrieval pipeline. Given the massive scale of video dataset, it is impractical to process all frames individually, as doing so would incur significant computational costs, excessive storage consumption, and reduced retrieval efficiency.
Therefore, we implement the TransNetV2 model \cite{ref_transnetv2} to segment each video into smaller shots, enabling us to extract only keyframes from each shot rather than processing the entire video. In the next stage, we employ an algorithm to select three representative keyframes from each shot. If the number of frames in a shot is not sufficient, we simply take all of them. Otherwise, we will choose 3 frames based on the percentiles [0.15, 0.5, 0.85]
\[
\text{keyframe}_i = \text{start} + \text{percentile}_i \times (\text{end} - \text{start}),
\]
where $\text{keyframe}_i$ denotes the $i$-th keyframe index, $\text{start}$ and $\text{end}$ are the indices of the first and last frames, and $\text{percentile}_i \in [0,1]$ is the normalized percentile position of the $i$-th keyframe.

\begin{figure*}[t]
    \centering
    \includegraphics[width=0.9\linewidth]{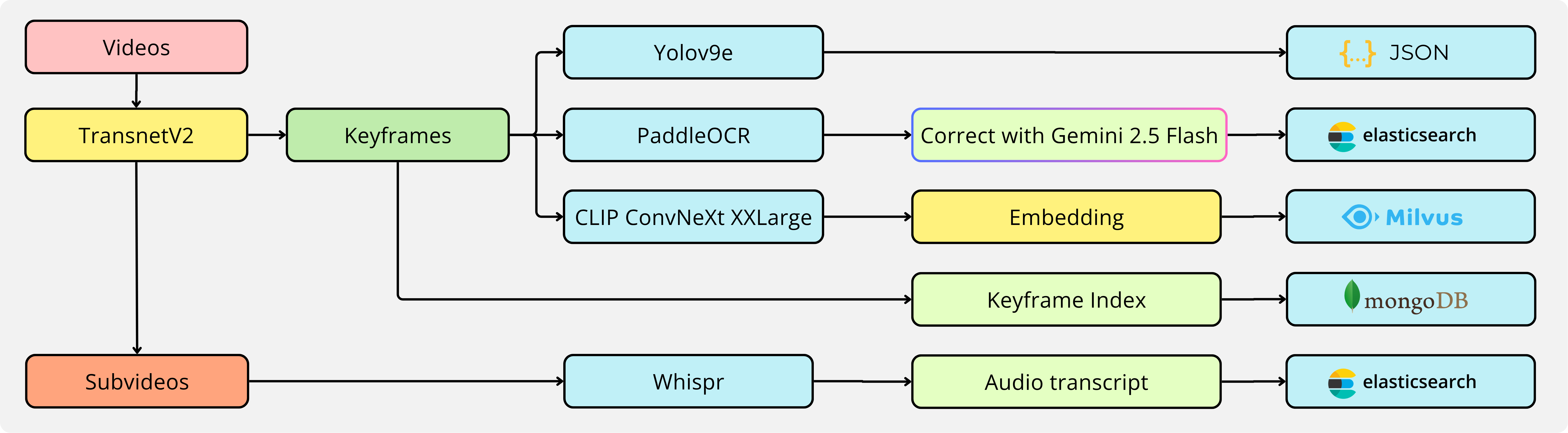}
    \caption{An overview of our fundamental multi-modal video retrieval system architecture.}
    \label{fig:system_overview}
    \includegraphics[width=0.9\linewidth]{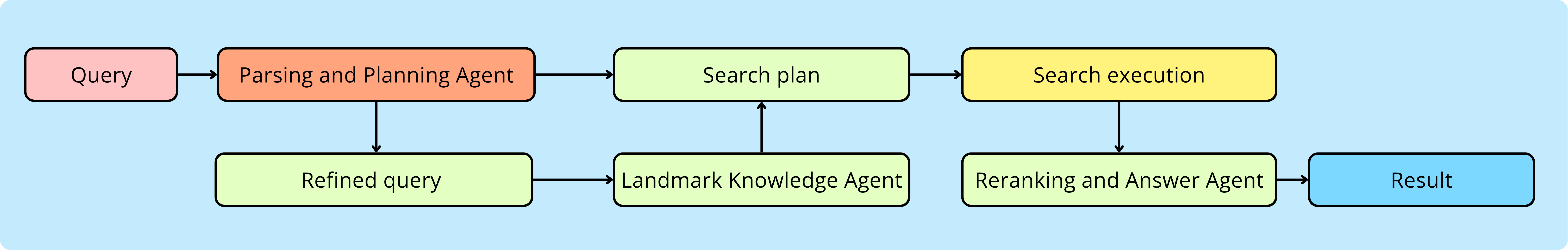}
    \caption{Overview of the LLandMark framework.}
    \label{fig:llandmark}
    \includegraphics[width=0.9\linewidth]{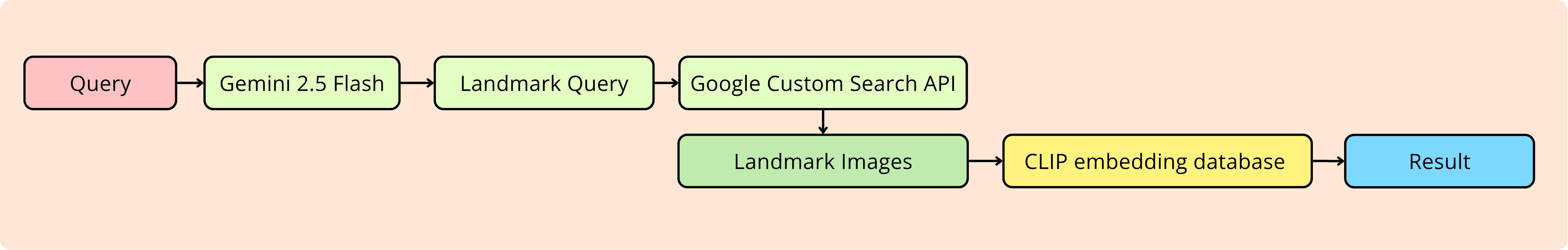}
    \caption{Workflow of LLM-Assisted Landmark Image-to-Image Retrieval}
    \label{fig:i2i}
\end{figure*}
\section{Methodology}
\subsection{Embedding-Based Search.}
After extracting keyframes and generating their corresponding indices, we represent all visual data within a vector space to enable efficient similarity-based retrieval. Two separate databases are maintained: one for keyframe embeddings and another for their structured identifiers.  Each keyframe index follows a hierarchical format of \textbf{group\_id/video\_id/frame\_id}.

This separation facilitates efficient filtering, sorting, and inclusion/exclusion operations. All index metadata are stored in MongoDB for rapid access and integration with higher-level retrieval modules.

For visual embedding, we employ the CLIP ConvNeXt-XXLarge (laion2B-s34B-b82K-augreg) model \cite{ref_clip,ref_openclip,ref_laionclip}, trained on the large-scale LAION-2B dataset. The encoded keyframe vectors are stored and indexed in Milvus, an open-source vector database optimized for large-scale similarity search, supporting fast cosine-based nearest-neighbor retrieval.

\subsection{Automatic Speech Recognition and Object-Based Retrieval.}

We employ WhisperX \cite{ref_whisperx2023} for automatic speech recognition, offering improved word alignment, speaker diarization, and robustness to accented or mixed-language speech. The resulting transcripts are indexed in Elasticsearch, where video segments are ranked using BM25 similarity and returned with highlighted context.

For object-based retrieval, we use the YOLOv9-e model \cite{ref_yolov9_2024} pre-trained on COCO. All keyframe detections are pre-computed and stored in a single JSON file with a lazy-loading mechanism for efficiency. Users can filter by target objects using AND/OR logic, and frames are ranked by matched object counts.
\begin{figure*}[t]
    \centering
    \scalebox{0.9}{
    \begin{minipage}[t]{0.49\textwidth}
        \centering
        \includegraphics[width=\linewidth]{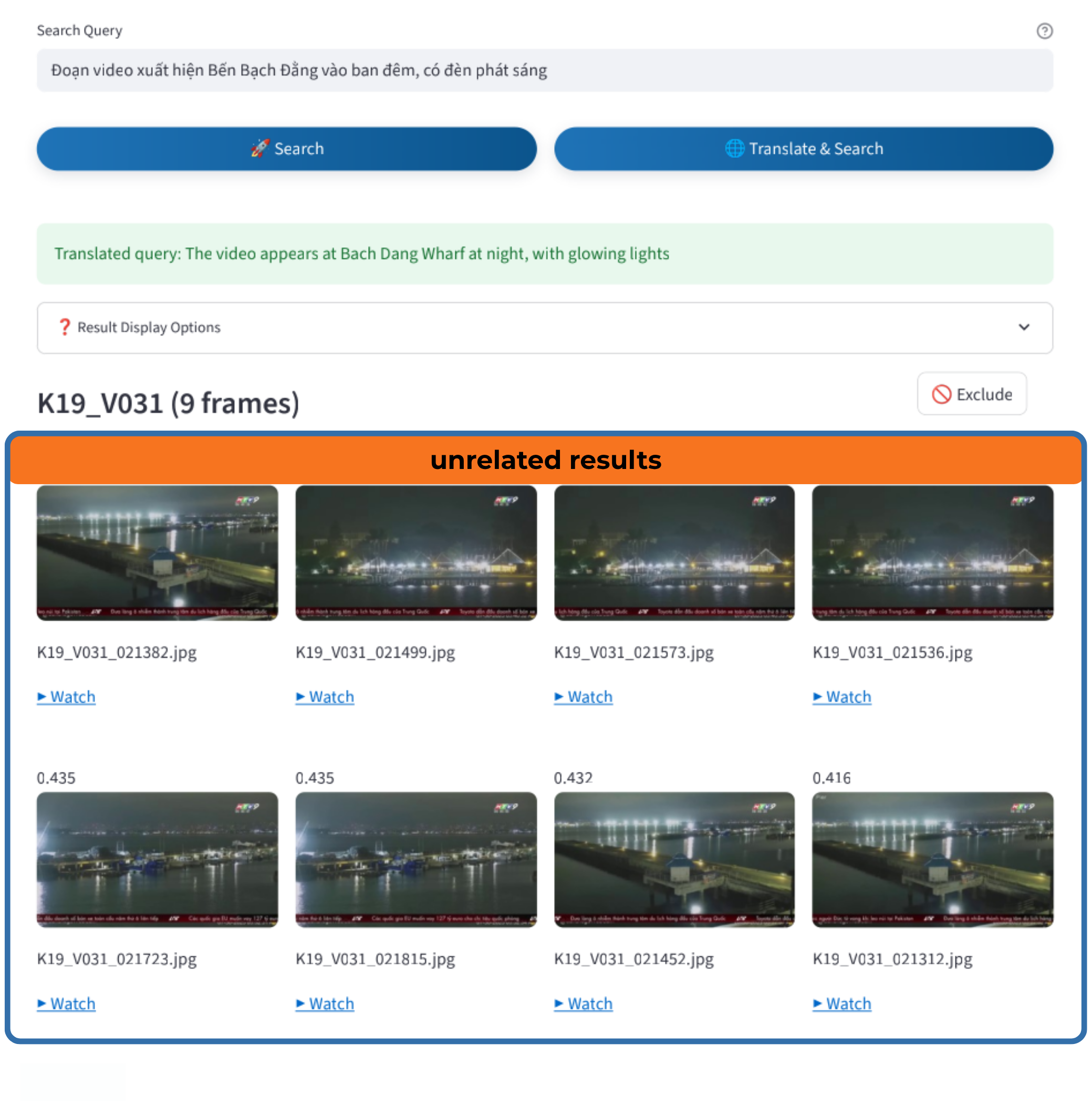}
        \caption{Embedding-based search}
        \label{fig:default_res}
    \end{minipage}
    \hfill
    \begin{minipage}[t]{0.48\textwidth}
        \centering
        \includegraphics[width=\linewidth]{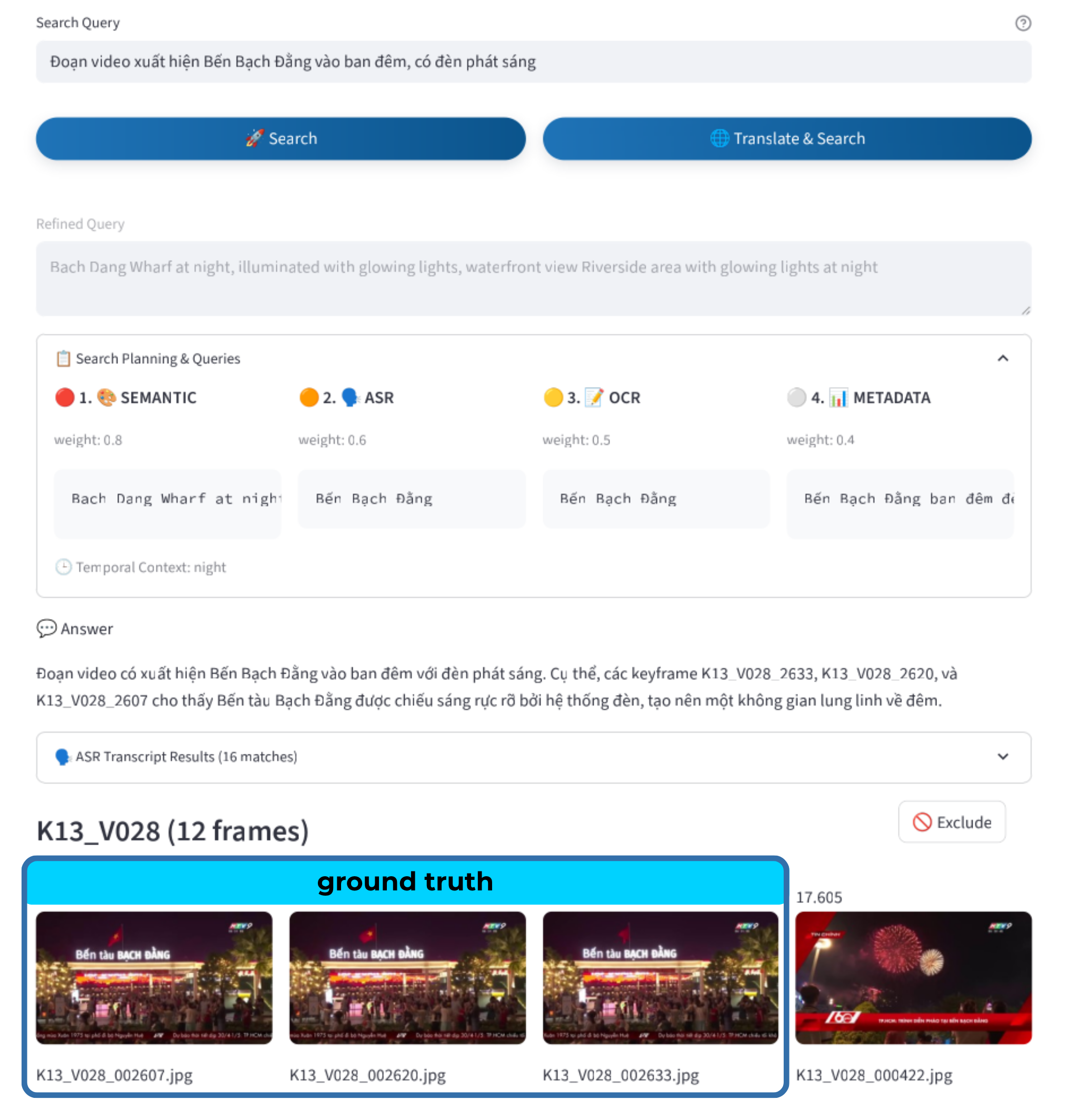}
        \caption{LLandMark search}
        \label{fig:llandmark_res}
    \end{minipage}
    }
\end{figure*}
\subsection{Optical Character Recognition with Gemini Refinement.}
We employed PaddleOCR \cite{ref_paddleocr} to extract textual content from video frames. While effective for Latin-based and East Asian scripts, it performed poorly on real-world Vietnamese text, \textbf{frequently misplacing or omitting diacritics} and thus altering meaning. To mitigate these errors, we first normalized outputs to \textbf{non-accent forms} to preserve lexical integrity and avoid propagation of corrupted tokens in later stages. To restore text quality, we introduced a \textbf{post-processing pipeline} based on the large language model Gemini 2.5 Flash, integrated via LlamaIndex \cite{ref_llamaindex}. The model automatically reconstructs Vietnamese diacritics, corrects spelling, and removes OCR noise using structured batch prompts, producing linguistically consistent text suitable for downstream retrieval and classification tasks. Refined texts are indexed in Elasticsearch, each entry linked to metadata such as group ID, video number, frame position, and confidence score, enabling efficient keyword-based retrieval across large-scale video datasets. Using the BM25 ranking model, the system retrieves and highlights relevant textual segments, providing contextual cues for interpretation.

Our method for temporal retrieval identifies videos that match an ordered sequence of $N$ textual queries, $Q = \{q_1, \dots, q_N\}$. First, for each query $q_i$, an independent vector search yields a set of relevant videos, $V_i$, where a video's step-specific score, $S_i(v)$, is the maximum similarity of its retrieved keyframes.

To ensure a video contains the entire sequence, we perform a set intersection across all steps to find candidate videos:
$$ V_{\text{candidate}} = \bigcap_{i=1}^{N} V_i $$
These candidates are then ranked by a conservative aggregation function that rewards consistent relevance. The final score is the minimum score a video achieved across all steps, ensuring that only videos that are strongly relevant to every part of the sequence are ranked highly.
$$ \text{Score}(v) = \min_{i=1, \dots, N} \{S_i(v)\} $$

Our fundamental system's architecture, illustrated in \textbf{Fig. \ref{fig:system_overview}}, is a multi-modal pipeline designed for comprehensive video analysis. The process begins with TransNetV2 segmenting videos into shots to extract keyframes. These keyframes are then processed in parallel to create multiple indices: CLIP embeddings are stored in Milvus for semantic search; text extracted by PaddleOCR and corrected by Gemini 2.5 Flash is indexed in Elasticsearch; and object detections from YOLOv9-e are saved as a JSON file. A primary keyframe index is kept in MongoDB to unify these components.

Concurrently, audio from corresponding subvideos is transcribed by WhisperX, and indexed in Elasticsearch. This integrated, multi-database approach enables a powerful hybrid search system, capable of combining semantic, textual, and object-based queries to address complex retrieval tasks efficiently.

\subsection{LLandMark Framework}
\begin{figure*}[t]
    \centering
    \scalebox{0.9}{
    \begin{minipage}[t]{0.49\textwidth}
        \centering
        \includegraphics[width=\linewidth]{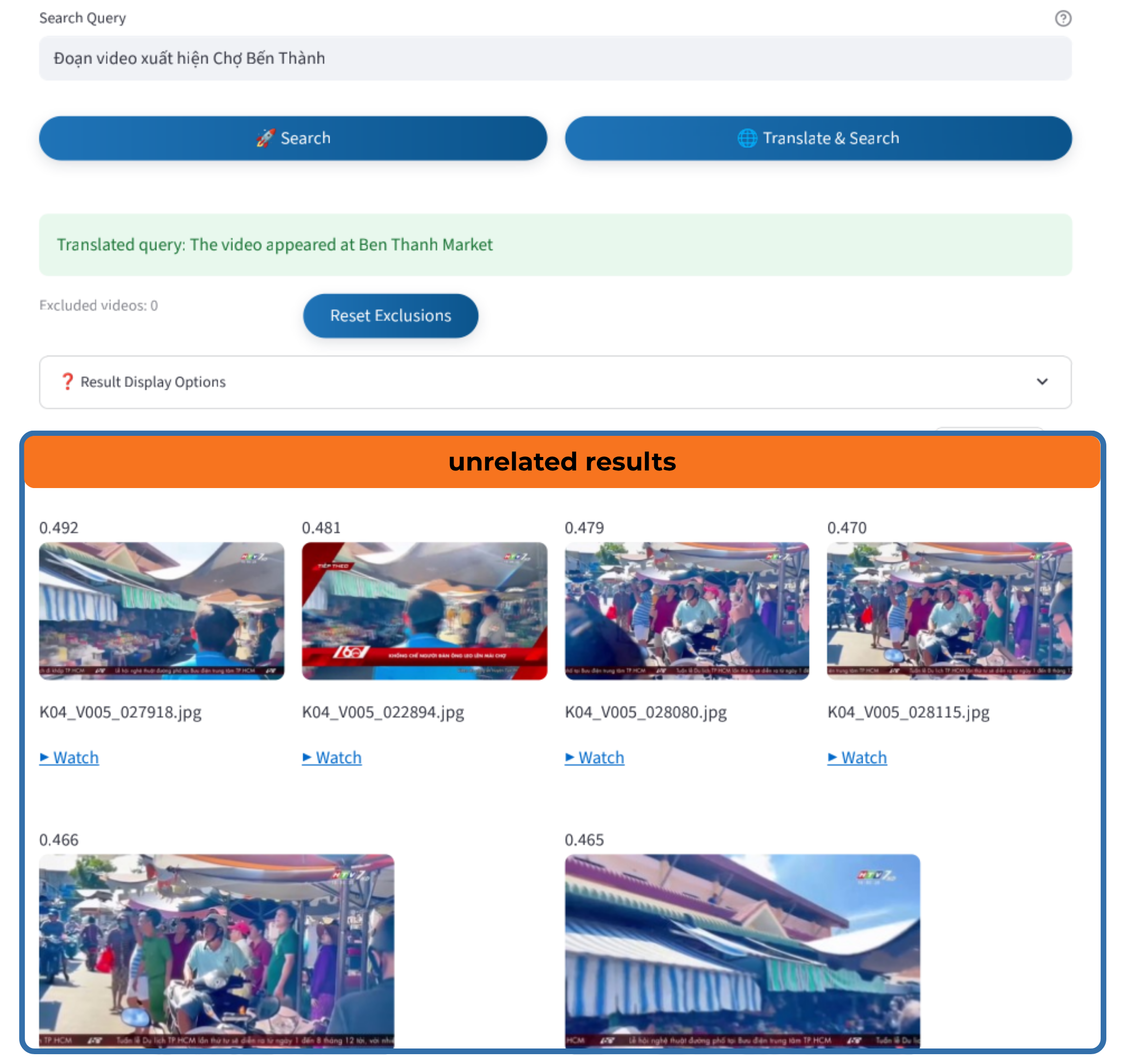}
        \caption{Embedding-based search}
        \label{fig:default_res1}
    \end{minipage}
    \hfill
    \begin{minipage}[t]{0.50\textwidth}
        \centering
        \includegraphics[width=\linewidth]{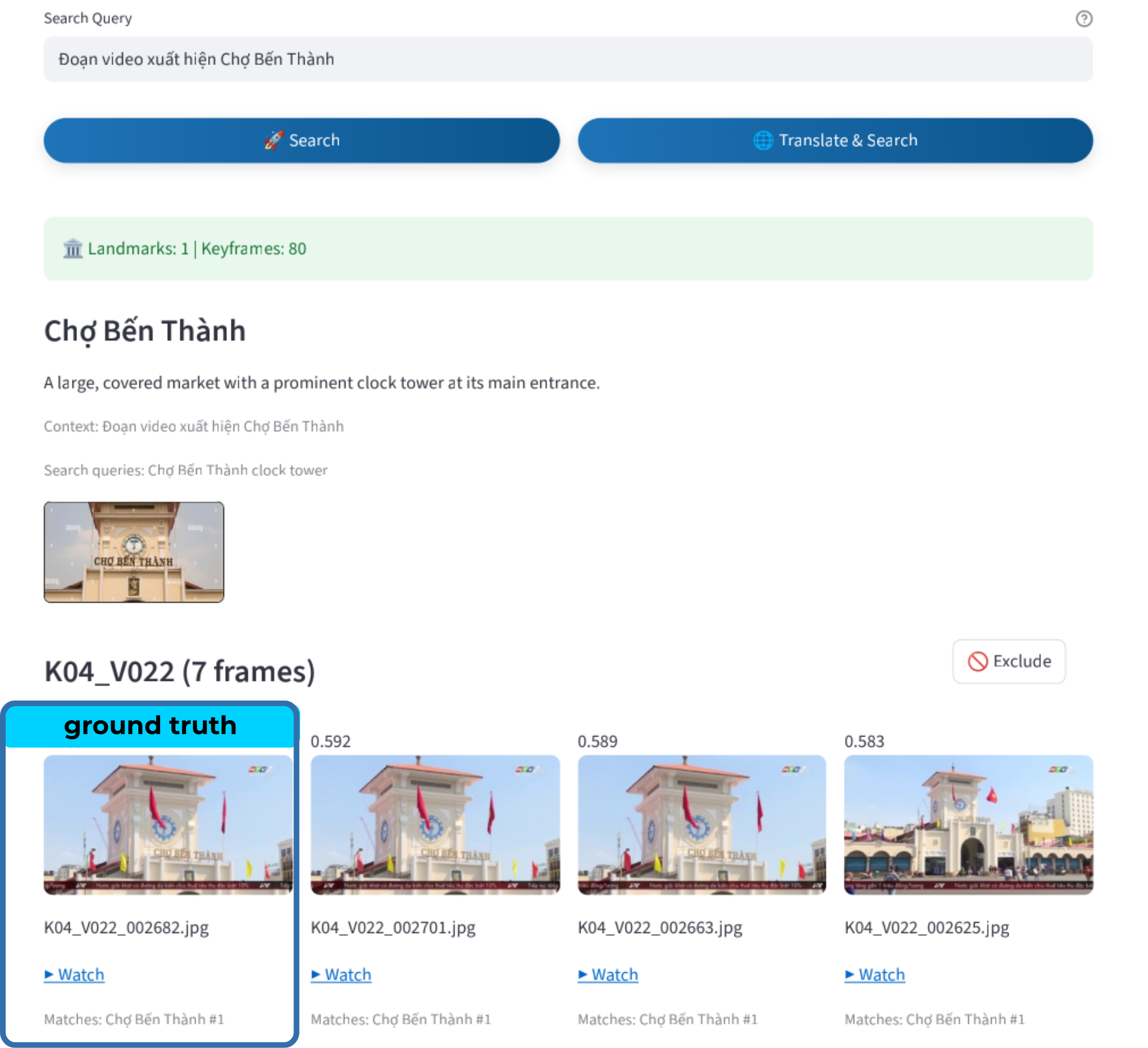}
        \caption{LLM-Assisted Landmark Image-to-Image search}
        \label{fig:i2i_res1}
    \end{minipage}
    }
\end{figure*}

The proposed LLandMark system is a modular multi-agent framework designed for adaptive, multimodal, and explainable video retrieval. As illustrated in \textbf{Fig.~\ref{fig:llandmark}}, it operates through four tightly integrated components: (1) query parsing and strategic planning, (2) landmark-specific knowledge enhancement, (3) parallel multimodal search, and (4) reranking and answer synthesis.

\textbf{Query Parsing and Planning Agent.} 
Upon receiving a user query in Vietnamese or English, the system first analyzes its intent and constructs a structured \texttt{SearchPlan} - a weighted set of search steps spanning semantic, ASR, OCR, and object modalities.  
For semantic search, queries are translated into descriptive English to maximize alignment with the CLIP embedding space. For ASR and OCR, landmark names and specific Vietnamese terms are retained verbatim to ensure exact matching against indexed transcripts and on-screen text. The system also detects landmark entities, flagging them for specialized enhancement in the next stage.

\textbf{Landmark Knowledge Enhancement.} 
When a landmark is detected, the Landmark Knowledge Agent enriches the \texttt{SearchPlan} using a curated knowledge base of Vietnamese landmarks, each described by detailed visual and architectural attributes. It replaces the landmark name in the semantic query with a rich descriptive prompt.  
For example, \textit{“St. Joseph's Cathedral”} becomes \textit{“Twin square bell towers, dark gray stone, Gothic architecture, neo-Gothic facade.”}  
This reformulation bridges the semantic gap in CLIP’s visual embedding space, enabling retrieval based on appearance rather than lexical identity.

\textbf{Parallel Multimodal Search.} 
The orchestrator executes the enhanced \texttt{SearchPlan} concurrently across all search modules:
\begin{itemize}
    \item \textbf{Semantic Search:} Descriptive English embeddings are used to search visually similar keyframes in Milvus.
    \item \textbf{ASR/OCR Search:} Vietnamese keywords query the Elasticsearch index for spoken or on-screen occurrences.
    \item \textbf{Object Filtering:} COCO-based detections filter keyframes using AND/OR logical conditions.
\end{itemize}

\textbf{Fusion and Answer Generation.} 
Retrieved results are combined by the Reranking and Answer Agent, which fuses multimodal scores using a weighted averaging formula:
\[
    \text{Score}_{\text{combined}} = 
    \frac{\sum_{m \in M} w_m \cdot s_m}{\sum_{m \in M} w_m},
\]
where $s_m$ and $w_m$ represent the score and weight of modality $m$.  
Top-ranked frames are grouped by video, and contextual evidence (images, ASR snippets, OCR text, detected objects) is packaged for a multimodal LLM, which synthesizes a coherent, grounded natural-language answer referencing specific video frames.

\subsection{LLM-Assisted Landmark Image-to-Image Retrieval}

To address the limits of text-based retrieval for landmark queries, we built a fully automated image-to-image search pipeline. It achieves higher visual fidelity by using real-world images as queries and removes the need for manual image input. As shown in \textbf{Fig.~\ref{fig:i2i}}, the workflow proceeds as follows:

\begin{enumerate}
    \item \textbf{Landmark Identification and Query Generation:} The agent detects landmark names and produces targeted, descriptive queries optimized for web image search (e.g., "The Imperial City of Hue from above").

    \item \textbf{Reference Image Acquisition:} These queries are executed through the Google Custom Search API to retrieve a small set of representative landmark images. 

    \item \textbf{Image-to-Image Similarity Search:} Retrieved images are encoded using the CLIP ConvNeXt-XXLarge image encoder. The resulting embeddings are used to search against keyframe embeddings stored in Milvus, producing visually similar candidates.

    \item \textbf{Result Aggregation and Reranking:} Candidate keyframes from all reference images are merged, duplicates are resolved by keeping the highest similarity score, and the final set is reranked by confidence.
\end{enumerate}

This pipeline reduces ambiguity in text-based queries by grounding retrieval in real landmark imagery, yielding more accurate and contextually aligned results.

\section{Experimental Results}

\begin{figure*}[t]
    \centering
    \includegraphics[width=0.9\linewidth]{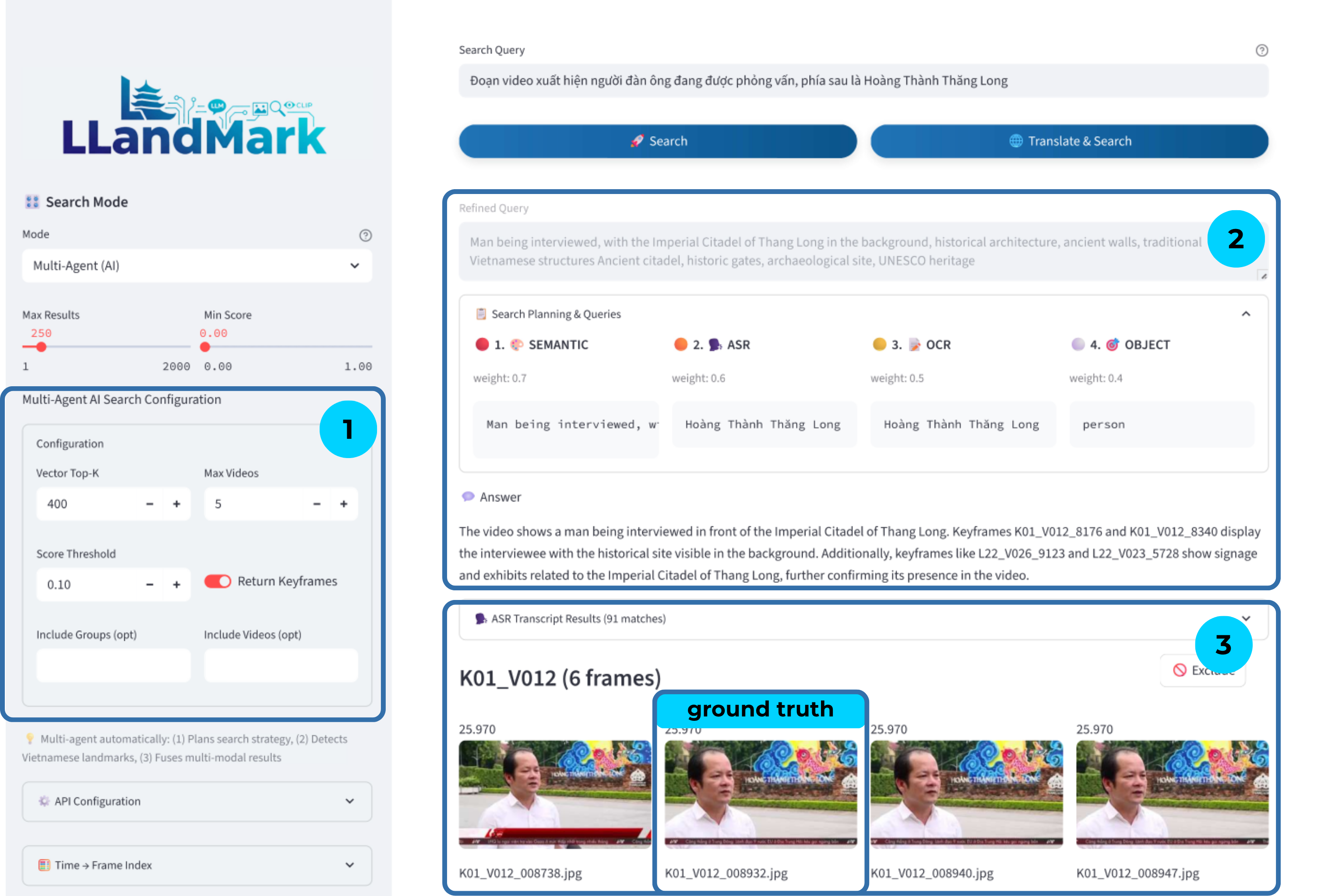}
    \caption{Illustration of the LLandMark search process.}
    \label{fig:detailllandmark}
\end{figure*}

\subsection{Experimental Setup}
Following the official Codabench evaluation protocol, system performance is measured using the \textbf{Mean of Top-k R-Scores}, averaged across $k \in \{1, 5, 20, 50, 100\}$. For each query, the R-Score quantifies the accuracy of a retrieved frame relative to the ground truth, depending on the task type:
\begin{itemize}
\item \textbf{Textual-KIS:} A result is correct if the predicted video matches the ground-truth video name and the retrieved frame index lies within the correct frame range.
\item \textbf{Visual QA:} In addition to frame alignment, the textual answer must exactly match the reference answer.
\item \textbf{Temporal Reasoning:} The R-Score is computed as the proportion of predicted frames overlapping with the ground-truth temporal segment, within a predefined tolerance window.
\end{itemize}
The final score for each query is computed as:
\[
\text{Final Score} = \frac{1}{5} \sum_{k \in \{1,5,20,50,100\}} \max_{1 \leq i \leq k} R\text{-Score}(r_i),
\]
where $r_i$ denotes the $i$-th ranked retrieval result. The overall system score is the mean of query scores across all benchmark subsets.
\subsection{Quantitative Results}
\begin{figure*}[t]
    \centering
    \includegraphics[width=0.9\linewidth]{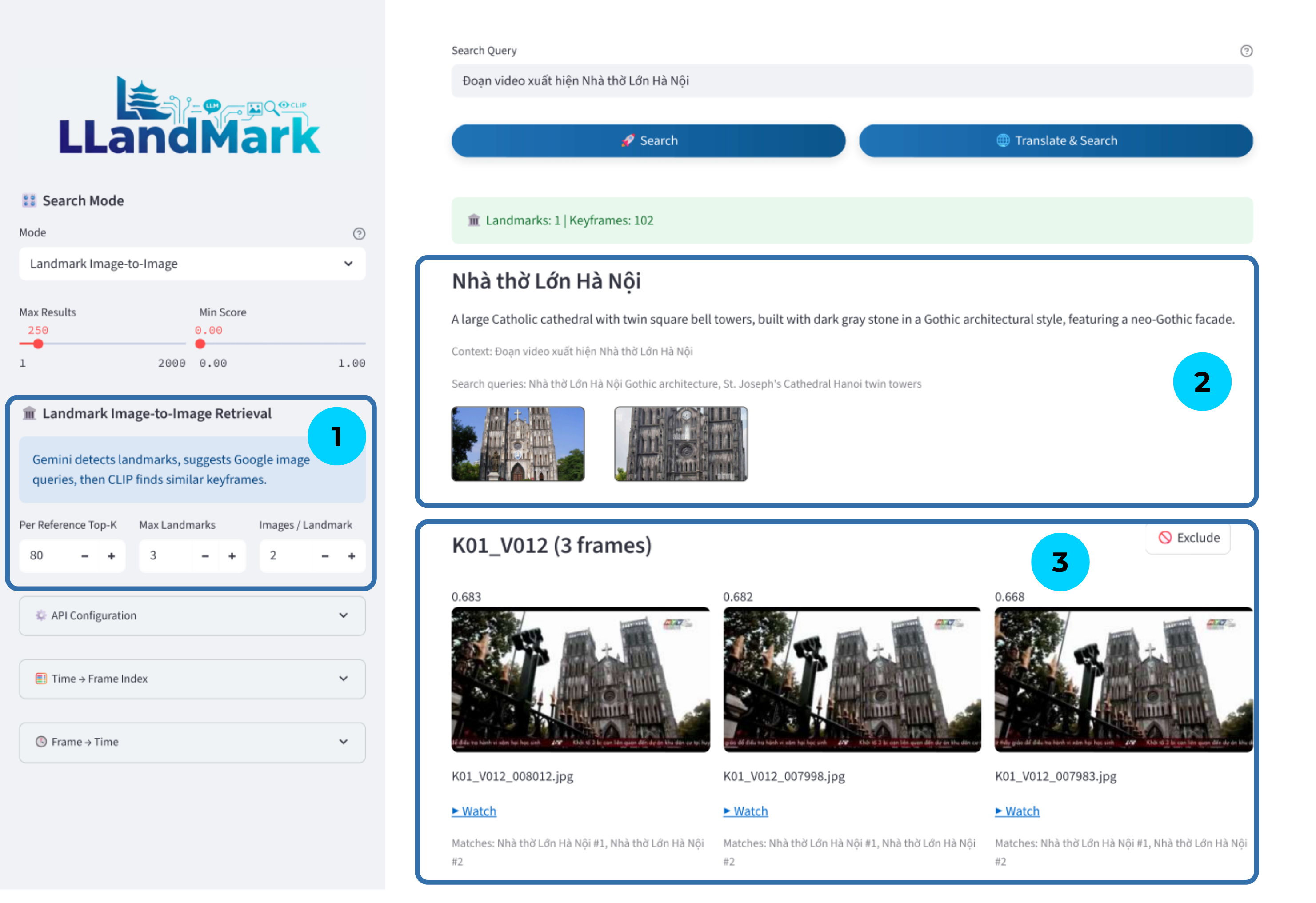}
    \caption{User interface for LLM-Assisted Landmark Image-to-Image feature}
    \label{fig:uii2i}
\end{figure*}
Table~\ref{tab:quantitative_results} summarizes the performance of the proposed \textbf{LLandMark} framework in the \textbf{qualifier round} of the HCMAIC~2025 challenge. The system achieved a total score of \textbf{77.40} out of a maximum of \textbf{88 points}, demonstrating competitive performance across progressively challenging evaluation rounds. Among more than \textbf{680 registered teams}, LLandMark ranked within the \textbf{top 56 teams} selected for official qualification, confirming its robustness and effectiveness in large-scale multimodal retrieval.

\begin{table}[h]
\centering
\caption{Performance across evaluation rounds in the HCMAIC~2025 qualifier.}
\label{tab:quantitative_results}
\begin{tabular}{lcccc}
\hline
\textbf{Round} & \textbf{Round 1} & \textbf{Round 2} & \textbf{Round 3} & \textbf{Total} \\ 
\hline
Score & 20.00 & 28.20 & 29.20 & \textbf{77.40} \\
Max-score & 23.00 & 30.00 & 35.00 & \textbf{88.00} \\
\hline
\end{tabular}
\end{table}

Table~\ref{tab:query_distribution} provides the distribution of benchmark queries across the three task categories-Keyword-based Information Search (KIS), Question Answering (QA), and Temporal Reasoning and Keyframe Extraction (TRAKE). The dataset complexity increases across rounds, with Round~3 containing the largest number of queries and serving as the most demanding evaluation phase.

\begin{table}[h]
\centering
\caption{Distribution of evaluation queries across tasks and rounds.}
\label{tab:query_distribution}
\begin{tabular}{lccc}
\hline
\textbf{Round} & \textbf{KIS} & \textbf{QA} & \textbf{TRAKE} \\ 
\hline
1 & 17 & 3 & 3 \\
2 & 26 & 2 & 2 \\
3 & 29 & 4 & 2 \\
\hline
\textbf{Total} & \textbf{72} & \textbf{9} & \textbf{7} \\
\hline
\end{tabular}
\end{table}

These results indicate that \textbf{LLandMark} maintains strong and consistent performance across all task types in the qualifier stage, effectively integrating multimodal reasoning, OCR correction, and landmark-aware search to achieve competitive retrieval accuracy at national scale.

\subsection{Qualitative Results}

To evaluate the effectiveness of our approach, we compare the retrieval results of our specialized modes with the embedding-based search.

As shown in \textbf{Fig.~\ref{fig:default_res}} and \textbf{\ref{fig:llandmark_res}}, the first query is "The video appears at Bach Dang Wharf at night, with glowing lights". The embedding-based search (\textbf{Fig.~\ref{fig:default_res}}) fails to resolve this complex, landmark-related query and returns unrelated results. In contrast, our LLandMark mode (\textbf{Fig.~\ref{fig:llandmark_res}}) successfully interprets the query, and retrieves the correct video segment.

Similarly, \textbf{Figures~\ref{fig:default_res1}} and \textbf{\ref{fig:i2i_res1}} illustrate the query "The clip shows Ben Thanh Market". The baseline CLIP model~\cite{ref_clip} (Fig.~\ref{fig:default_res1}) misinterprets the landmark as a generic market, while our LLM-Assisted Landmark Image-to-Image mode (Fig.~\ref{fig:i2i_res1}) accurately identifies “Ben Thanh Market”, retrieves its reference images, and performs image-to-image matching to return the correct video as the top-ranked result.

\section{System Usage}
\subsection{Overall User Interface}
The system’s user interface provides an efficient environment for multimodal video retrieval, organized into a left panel for configuration and a right panel for query input and result visualization. To support Vietnamese users and maintain semantic consistency, an automatic translation tool converts Vietnamese queries into English for improved CLIP embedding alignment. The UI offers multiple retrieval modes, letting users switch between semantic, OCR, ASR, and object-based search, with an include/exclude mechanism for fine-grained filtering. An integrated index-to-time calculator maps retrieved keyframes to their original timestamps for direct video inspection. These components make the interface adaptable, transparent, and accessible while supporting detailed multimodal analysis.

\subsection{LLandMark}

LLandMark provides an interactive interface for iterative query refinement, multimodal visualization, and real-time feedback. Users can observe how agents interpret, plan, and synthesize retrieval results, ensuring transparent and explainable behavior.

\textbf{Fig.~\ref{fig:detailllandmark}} shows the main interface. Section \textbf{(1)} handles key parameters such as retrieval thresholds and keyframe settings. For a query like “The video clip shows a man being interviewed, with the Thang Long Imperial Citadel in the background,” results appear in section \textbf{(3)}.

In section \textbf{(2)}, the \textit{Refined Query} panel shows how the system reformulates and clarifies landmark descriptions, while the \textit{Answer} panel presents ranked results from the Reranking and Answer Agent. \textit{Search Planning and Queries} panel visualizes dynamic weighting among the four modalities: Semantic, ASR, OCR, Object, along with the ASR transcript.

Retrieved keyframes appear in a structured grid with OCR text, semantic labels, detected objects, and confidence scores, along with corresponding ASR transcripts. This layout makes the reasoning process interpretable and allows users to trace the evidence that leads to the final output.

\subsection{LLM-Assisted Landmark Image-to-Image}

As shown in \textbf{Fig.~\ref{fig:uii2i}}, the LLM-Assisted Landmark Image-to-Image mode provides an interface for controlling and interpreting the retrieval process. Panel \textbf{(1)} lets users configure retrieval settings such as “Per Reference Top-K,” “Max Landmarks,” and “Images/Landmark.” For a query like \textit{"The video shows St. Joseph’s Cathedral in Hanoi"}, the system’s interpretation appears in panel \textbf{(2)}, confirming the landmark, summarizing it, and displaying the exact search queries used to retrieve web reference images. These images are then used as visual queries for CLIP-based similarity search. Final keyframes appear in \textbf{(3)}, grouped by source video.


\section{Conclusion}

LLandMark is a modular multi-agent framework for adaptive, multimodal, and explainable video retrieval, combining query parsing, knowledge enhancement, parallel multimodal search, and LLM-based synthesis to handle complex natural-language queries. Its OCR refinement pipeline improves Vietnamese text recognition by restoring diacritics and correcting noise, while the LLM-assisted image-to-image module grounds landmark queries in visual references to enhance precision and contextual relevance. Together, these components demonstrate how structured retrieval planning augmented with LLM reasoning enables interpretable, flexible, and culturally aware multimedia search, forming a foundation for future systems that integrate vision, language, and reasoning for human-centered video retrieval.
\bibliography{aaai2026}


\end{document}